\documentclass[sigconf,nonacm]{aamas} 

\usepackage{balance} 
\usepackage{array}
\usepackage{code_formatting}
\usepackage{courier}

\usepackage{tikz}

\acmSubmissionID{???}

\title{Multiplayer Support for the Arcade Learning Environment}

\author{J K. Terry}\authornote{Equal contribution}
\affiliation{
  \institution{Swarm Labs}
  \institution{University of Maryland, College Park}
  }
\email{j.k.terry@swarmlabs.com}

\author{Benjamin Black}\authornotemark[1]
\affiliation{
  \institution{Swarm Labs}
  \institution{University of Maryland, College Park}
  }
\email{benjamin.black@swarmlabs.com}

\author{Luis Santos}
\affiliation{
  \institution{Swarm Labs}
  \institution{University of Maryland, College Park}
  }
\email{luis.santos@swarmlabs.com}

\begin{abstract}
  The Arcade Learning Environment (``ALE'') is a widely used library in the reinforcement learning community that allows easy programmatic interfacing with Atari 2600 games, via the Stella emulator. We introduce a publicly available extension to the ALE that extends its support to multiplayer games and game modes. This interface is additionally integrated with PettingZoo to allow for a simple Gym-like interface in Python to interact with these games. We additionally introduce experimental baselines for all environments included.
\end{abstract}


\newcommand{\BibTeX}{\rm B\kern-.05em{\sc i\kern-.025em b}\kern-.08em\TeX}

\begin{document}


\pagestyle{fancy}
\fancyhead{}


\maketitle 


\section{Introduction}

\begin{figure}
    \centering
    \begin{minipage}{.48\linewidth}
    \centering
    \includegraphics[scale=0.55]{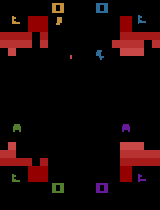}
    \end{minipage}
    \begin{minipage}{.48\linewidth}
    \centering
    \includegraphics[scale=0.55]{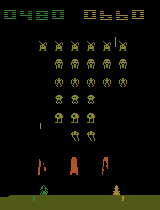}
    \end{minipage}
    \caption{Warlords (left), and Space Invaders (right) are two of the more visually interesting games included. In Warlords, 4 players control paddles to deflect the ball away from the bricks and their colored "warlord" in their corner. Players die when the ball hits their warlords, and the last player standing wins. In the multiplayer mode of space Invaders, both players control ships that can move and shoot at waves of enemies that shoot at them. Players are rewarded for killing enemies, and the game ends when the players have lost three combined lives. We additionally include experimental baselines for all new environments.}
    \label{fig:my_label}
\end{figure}

The ALE was first introduced in \citet{bellemare2013arcade}, with further improvements introduced in \citet{machado2018revisiting} (for single player games only). It was a fork of the Stella Atari 2600 emulator, that allowed for a simple C interface to take actions in a supported Atari game, and receive the resulting screen state and game reward.

Reinforcement Learning (``RL'') considers learning a policy --- a function that takes in an observation from an environment and emits an action --- that achieves the maximum expected discounted reward when playing in an environment. Atari 2600 games were viewed as a useful challenge to pursue for reinforcement learning as they were interesting and difficult to humans, and contained a wide array of premade games to test an AI against to prevent researcher bias and to allow for comprehensive benchmarking of a technique \citep{machado2018revisiting}. They were also special for having visual observation spaces, and were practically useful because they had small graphical observation spaces, ran fast, had generally simple game mechanics that were still challenging to humans, and had a frequently updating well defined reward.

The ALE served as the benchmark suite in \citet{mnih2013playing} and \citet{mnih2015human}, which introduced and showed the power of the Deep Q Network method of learning games, essentially creating the field of Deep Reinforcement Learning in the process. These games have remained a very popular benchmark in deep reinforcement learning, as developing very effective methods for all games is still a difficult challenge years later \citep{hessel2018rainbow, ecoffet2019go}. These ALE games were later packaged with a much simpler and more general Python interface into Gym \citep{brockman2016openai}, which is how the ALE environments are most frequently used today.

Following the boom in single agent Deep Reinforcement Learning research, work has been ongoing to learn optimal policies for agents in environments where multiple agents interact, be the interactions strictly competitive, strictly cooperative, or mixed sum \citep{silver2017mastering, lowe2017multi, gupta2017cooperative, TerryParameterSharing}. For the same reasons that single player Atari 2600 games became very popular suite of benchmark environments, we feel that the multiplayer Atari 2600 games are uniquely well suited benchmark environments for multi-agent reinforcement learning, and have extended the ALE to support them, allowing easy programmatic interfacing with multiplayer Atari 2600 games for the first time, and a comprehensive suite of differing multi-agent RL benchmarks for the first time.

\section{API}

In order to support multiplayer games, a few changes to the existing ALE API had to be made, outlined in \citet{bellemare2013arcade}. The \texttt{getAvailableModes} method was overloaded to receive number of players as an optional argument and return game modes with that number of players. The \texttt{step} method was overloaded to accept a list of actions, and return a list of rewards (one reward and action for each player). The \texttt{allLives} method was added to return the number of lives for each player as a list. We used the standard of 0 lives meaning the agent is alive, and the game will end after one more life. All these changes are backwards compatible and will not impact the single player modes of a game, except for one: we changed game modes to be more aligned with those listed in their manuals to make adding multiplayer modes feasible.

An example of using ALE API for the multiplayer mode of \emph{space invaders} is shown below:

\vspace{15pt}
\begin{Verbatim}[commandchars=\\\{\},frame=lines,
               framesep=2mm]
\PY{n}{ale} \PY{o}{=} \PY{n}{multi\PYZus{}agent\PYZus{}ale\PYZus{}py}\PY{o}{.}\PY{n}{ALEInterface}\PY{p}{(}\PY{p}{)}
\PY{n}{ale}\PY{o}{.}\PY{n}{loadROM}\PY{p}{(}\PY{l+s+s2}{\PYZdq{}}\PY{l+s+s2}{ROMS/space\PYZus{}invaders.bin}\PY{l+s+s2}{\PYZdq{}}\PY{p}{)}
\PY{n}{minimal\PYZus{}actions} \PY{o}{=} \PY{n}{ale}\PY{o}{.}\PY{n}{getMinimalActionSet}\PY{p}{(}\PY{p}{)}
\PY{n}{modes} \PY{o}{=} \PY{n}{ale}\PY{o}{.}\PY{n}{getAvailableModes}\PY{p}{(}\PY{n}{num\PYZus{}players}\PY{o}{=}\PY{l+m+mi}{2}\PY{p}{)}
\PY{n}{ale}\PY{o}{.}\PY{n}{setMode}\PY{p}{(}\PY{n}{modes}\PY{p}{[}\PY{l+m+mi}{0}\PY{p}{]}\PY{p}{)}
\PY{n}{ale}\PY{o}{.}\PY{n}{reset\PYZus{}game}\PY{p}{(}\PY{p}{)}
\PY{k}{while} \PY{o+ow}{not} \PY{n}{ale}\PY{o}{.}\PY{n}{game\PYZus{}over}\PY{p}{(}\PY{p}{)}\PY{p}{:}
    \PY{n}{observation} \PY{o}{=} \PY{n}{ale}\PY{o}{.}\PY{n}{getScreenRGB}\PY{p}{(}\PY{p}{)}
    \PY{n}{a1} \PY{o}{=} \PY{n}{policy1}\PY{p}{(}\PY{n}{observation}\PY{p}{)}
    \PY{n}{a2} \PY{o}{=} \PY{n}{policy2}\PY{p}{(}\PY{n}{observation}\PY{p}{)}
    \PY{n}{r1}\PY{p}{,}\PY{n}{r2} \PY{o}{=} \PY{n}{ale}\PY{o}{.}\PY{n}{act}\PY{p}{(}\PY{p}{[}\PY{n}{a1}\PY{p}{,}\PY{n}{a2}\PY{p}{]}\PY{p}{)}
\end{Verbatim}

\begin{table*}
\caption{All Multiplayer ROMs and Game Types Supported}
\label{table:game_support}
\begin{minipage}[t]{0.48\linewidth}
\begin{tabular}[t]{m{2cm}m{1.1cm}ll}
    ROM (Modes) & Image & Game theory & Players  \\ \midrule
    Boxing   & \vspace{3pt}\includegraphics[scale=0.2]{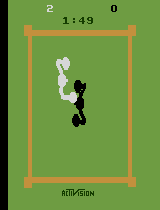}  & Competitive & 2  \\ \hline
    Combat (Tank, Plane) & \vspace{3pt}\includegraphics[scale=0.2]{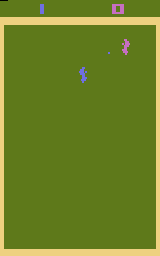}  & Competitive & 2  \\ \hline
    Double Dunk & \vspace{3pt}\includegraphics[scale=0.2]{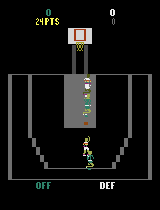}  & Competitive & 2  \\ \hline
    Entombed (Cooperative, Competitive) & \vspace{3pt}\includegraphics[scale=0.2]{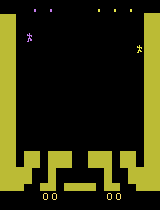} & \begin{tabular}{c}Competitive  \\ \& \\  Cooperative\end{tabular}   & 2  \\ \hline
    Flag Capture  & \vspace{3pt}\includegraphics[scale=0.2]{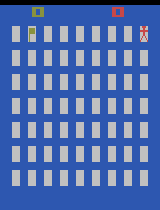} & Competitive & 2  \\ \hline
    Ice Hockey & \vspace{3pt}\includegraphics[scale=0.2]{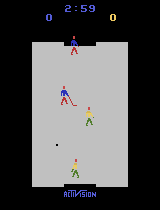}  & Competitive & 2  \\ \hline
    Joust & \vspace{3pt}\includegraphics[scale=0.2]{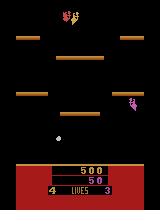}  & Mixed & 2  \\ \hline
    Mario Bros & \vspace{3pt}\includegraphics[scale=0.2]{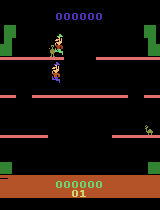}  & Mixed & 2  \\ \hline
    Maze Craze & \vspace{3pt}\includegraphics[scale=0.2]{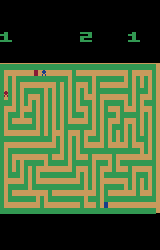}  & Competitive & 2  \\ \hline
\end{tabular}
\end{minipage}
\begin{minipage}[t]{0.48\linewidth}
\begin{tabular}[t]{m{2cm}m{1.1cm}ll}
    ROM (Modes) & Image & Game theory & Players  \\ \midrule
    Othello  & \vspace{3pt}\includegraphics[scale=0.2]{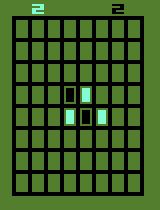} & Competitive & 2  \\ \hline
    Video Olympics (Pong, Basketball, Foozpong, Quadrapong, Volleyball)  & \vspace{3pt}\includegraphics[scale=0.2]{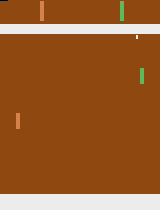} & Competitive & 2/4  \\ \hline
    Space Invaders & \vspace{3pt}\includegraphics[scale=0.2]{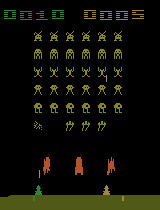}  & Mixed & 2  \\ \hline
    Space War & \vspace{3pt}\includegraphics[scale=0.2]{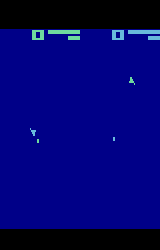}  & Competitive & 2  \\ \hline
    Surround & \vspace{3pt}\includegraphics[scale=0.2]{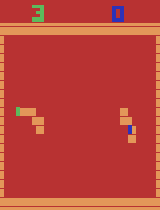}  & Competitive & 2  \\ \hline
    Tennis & \vspace{3pt}\includegraphics[scale=0.2]{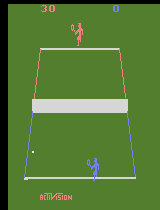}  & Competitive & 2  \\ \hline
    Video Checkers & \vspace{3pt}\includegraphics[scale=0.2]{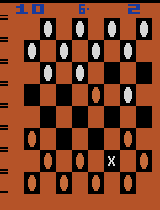}  & Competitive & 2  \\ \hline
    Warlords & \vspace{3pt}\includegraphics[scale=0.2]{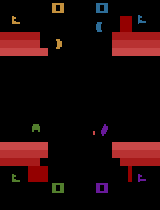}  & Competitive & 4  \\ \hline
    Wizard of Wor & \vspace{3pt}\includegraphics[scale=0.2]{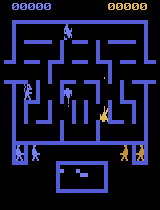}  & Mixed & 2  \\ \hline
\end{tabular}
\end{minipage}

\flushleft
\vspace{10pt}

\vspace{2cm}
\end{table*}

\begin{figure*}[h!]
    \centering
    \scalebox{0.8}{\input{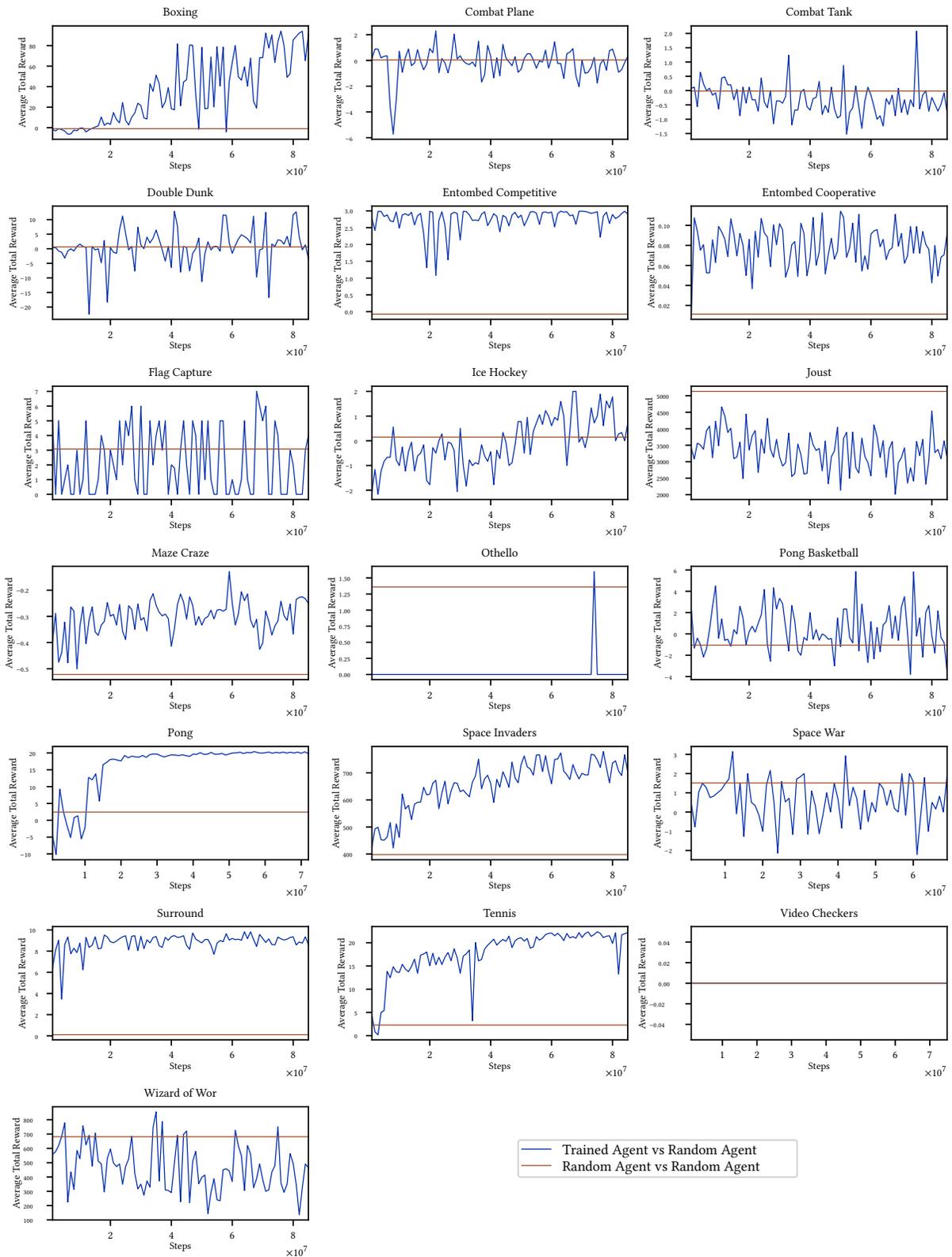}}
    \caption{Average total reward per step for policy trained with ApeX DQN on several Atari environments.}
    \label{fig:atari_results}
\end{figure*}

\section{Games Included}
\label{section:games_included}

We added support for 18 multiplayer games, 14 of which were extended from existing ALE games, and 4 are brand new. All added games are shown in Table \ref{table:game_support}. A few have alternate game modes which offer completely different game play, effectively adding 6 more games for a total of 24. These modes are listed as parenthesized names in Table \ref{table:game_support}. These modes as well as other noteworthy game modes are described in Appendix \ref{appendix:extended_modes}.  

Having a comprehensive and ``unbiased'' suite of games is considered key to the success of the ALE \citep{bellemare2013arcade, machado2018revisiting}. Of the games that were already in the ALE, we tried to add support as indiscriminately as possible, choosing 14/19 possible multiplayer games. The 4 we did not add multiplayer support to were Hangman (an asymmetry in the game play lets one player be guaranteed to win), Lost Luggage (the two different players are heterogeneous in ways that make adding this game beyond the scope of this work), as well as Freeway and Blackjack (they offer no interactions between the two players). Out of the four new games, we selected Combat, Joust and Warlords for their fame amongst human players and Maze Craze because of the very unique planning challenges it offers. All but Joust are multiplayer only games (and therefore couldn't have been included in the ALE before).

This selection process gave rise to an interesting mix of games that can be roughly categorized in 6 groups:

\textbf{1v1 Tournament}: Boxing, Combat, Ice Hockey, Video Olympics (two player), Double Dunk, Tennis and Space War. These games are zero-sum competitions, so a match or tournament needs to be run in order to compare the quality of different policies. They are also fast paced, with non-sparse reward in random play, and relatively  short dependencies between actions and rewards. But to master these games, your agent requires careful analysis of your opponent's position, prediction of their future movements, and very precise movements with quick reaction time. 

\textbf{Mixed-sum survival}: Joust, Mario Bros, Wizard of Wor and Space Invaders. These are 2 player games characterized by waves of dangerous computer-controlled opponents which players are rewarded for destroying. The computer-controlled opponents are the main source of risk and reward (allowing players to work together), but your opponent can still benefit by doing you harm. In Joust and Wizard of Wor, you can score points by directly destroying your opponent, in Space Invaders you score points when the Aliens destroy your opponent, and in Mario Bros your opponent can steal the reward you worked for. In all of these games, a robust agent has to be able to handle both aggressive competitive strategies and cooperative strategies from an opponent to maximize their score.

\textbf{Competitive racing}: Flag Capture, Entombed (Competitive Mode) and Maze Craze. These are fully competitive games more focused on understanding the environment than on the opponent to complete the race first. However as the agents develop, competitive strategies are possible. In Flag Capture, the information one player reveals about the map can be exploited by the other player. In Entombed, players have their ability to block their opponent using power-ups. In Maze Craze, we support numerous multiplayer modes (detailed in Appendix \ref{appendix:maze_craze}). These typically have little player interaction, but in three modes players can add fake walls that can be moved through, but can be confusing to the opponent.

\textbf{Long term strategy games}: Video checkers and Othello. These games are classic zero-sum 1v1 board games of long term strategy with an Atari interface. Due to this inefficient interface, a single turn in the board game will require many steps in the arcade game. In addition, like in the classic games, strong players must consider the consequences of moves many turns later, so very long term planning and credit assignment is required to create strong policies.

\textbf{Four-player free-for-all}: Warlords. Each player's objective is to be the last one standing out of the initial four players. Because of this, cooperation between players is possible until there are only 2 players left. For example, players can coordinate attacks against the strongest or best positioned player so that player will not have a large advantage at the end of the game when it is just 1v1.

\textbf{2v2 Tournament}: Video Olympics. The Video Olympics ROM includes 2v2 team versions of every 1v1 game type (specific modes are detailed in Appendix \ref{appendix:video_olympics}). The agents on one team will need to cooperate in order to both defend themselves and effectively attack the opposing team.

\textbf{Cooperative games}: Entombed (Cooperative Mode). In Entombed, the players are rewarded based on how long they can both stay alive in the environment. They have the ability to help each other out by using their power-ups to clear obstacles in the maze. This mode is detailed in Appendix \ref{appendix:entombed}.

Finally, we had to change the reward structure for games where one player can indefinitely stall the game by not acting (Double Dunk, Othello, Tennis and Video Checkers). We accomplished this by making a player lose the game with the associated negative reward if a player stalls the game for several hundred frames.

\section{Baselines}
In order to characterize the difficulty of these environments, and to provide a baseline for future work, we trained agents in every environment using self play/parameter sharing with the Ape-X DQN method \citep{Horgan2018APEX}, using the hyperparameters specified in \autoref{appendix:hyperparameters}. This choice was motivated by the experimental performance of Ape-X DQN in similar scenarios in \citep{TerryParameterSharing} and \citep{bard2019hanabi}. Training was done using RLlib \citep{liang2018rllib}, based on the PettingZoo versions of the environments in \citet{pettingZoo2020}, using standard Atari preprocessing wrappers from \citet{SuperSuit2020} described in \autoref{appendix:hyperparameters}.

To evaluate the quality of the trained agent as training progressed, we evaluate the total average reward the first (trained) agent received when playing against a random agent. The baseline is the reward of a random agent playing against a random agent. Choosing to evaluate the trained policy against a fixed policy which the agent was not trained against (the random policy) was an intentional choice to capture how well the self-play training generalized to a novel agent. The results of this evaluation are presented in Figure \ref{fig:atari_results}.

Several environment policies learned interesting emergent behavior. For example, in Boxing the agent learns to back up when its opponent punches and it learns to push its opponent into the wall so it cannot dodge as effectively. In other environments, the agent learns a very simple, but effective strategy. In Tennis, for example, the agent quickly moves into the optimal spot to receive a basic serve, and does not move from there. Surprisingly, in Joust and Wizard of Wor, the trained agent performed significantly worse than a random agent. One likely explanation is that this pathological training is the result of the mixed game theoretic structure of these two environments (discussed in \autoref{section:games_included}). A policy may end up focusing on combating the opponent, making both agents choose conservative policies, lowering the overall accumulation of reward. It is likely that a league system, similar to \citet{vinyals2019grandmaster}, is will be needed to overcome many of the challenges faced in learning these environments.

Animated gifs of the trained agent playing a random agent are included for all environments in the supplementary materials.

\section{Conclusion}

This work introduces multiplayer game support for the Arcade Learning Environment (ALE) for 18 ROMs, enabling 24 diverse multiplayer games. This builds off both the ubiquity and utility of Atari games as benchmarking environments for reinforcement learning, and the recent rise in research in multi-agent reinforcement learning.

We hope that this framework will enable accurate benchmarking for more general multi-agent reinforcement learning methods that can handle graphical observations, highly diverse game-play, and diverse multi-agent reward structures. 
We believe that multi-agent algorithms with these characteristics will generalize better to real world scenarios, and that no existing set of benchmarks has satisfied these needs particularly well.

The package is publicly available on PyPI and can be installed via \texttt{pip install multi-agent-ale-py}. AutoROM, a separate PyPI package, can be used to easily install the needed Atari ROMs in an automated manner. The Atari games here are additionally included with a simpler Python API in PettingZoo, which is akin to a multi-agent version of OpenAI's Gym library.

In order to better characterize the utility of each environment, we additionally have shown experimental self-play baselines for all new environments. These baselines indicate which environments have sparse reward, which ones can be trained easily with self-play, and which ones have pathological training under self-play. Future work can use these results to guide the direction of their research. These results will also allow the general community know what sort of results are trivial, and what sort of results indicate an interesting new method. It is likely that the usage of league, similar to \citet{vinyals2019grandmaster} will be required.




\bibliographystyle{ACM-Reference-Format} 
\bibliography{sample}

\clearpage

\appendix





\section{Extended Modes Documentation}
\label{appendix:extended_modes}

Many games have non-default modes that fundamentally change the nature of the game. We elaborate on one's we find notable, or treat as independent games, below.

\subsection{Combat}

Combat has two main style of play. Tank mode, where the player crawls around a field (potentially with obstacles), and plane mode, where the player cannot control their speed, only their direction. Within the tank category, there are a few options of interest. Table \ref{table:combat_modes} lists the game modes of interest

\begin{itemize}
    \item Maze: Map has obstacles which block movement and bullets
    \item Billiards: If this option is off, bullets are guided (turn when your tank turns). If it is on, bullets can and must (as in billiards) bounce off walls, allowing you to hit around corners. Note that you cannot hit yourself with a ricocheting bullet.
    \item Invisible: Tanks are not visible except when firing and when running into an obstacle. 
\end{itemize}

\begin{table}[h!]
\centering
\caption{Combat Tank modes}
\label{table:combat_modes}
\begin{tabular}{c|ccc}
    Mode & Maze & Billiards & Invisible \\ \midrule
    1 & F & F & F \\
    2 & T & F & F \\
    8 & F & T & F \\
    9 & T & T & F \\
    10 & F & F & T \\
    11 & T & F & T \\
    13 & F & T & T \\
    14 & T & T & T \\
\end{tabular}
\end{table}

In plane mode, there are two types of planes (Jet or Bi-Plane), which can either have strait or guided missiles. Jets are faster, and guided missile's direction can be changed by the palyer changing theier direction. Table \ref{table:plane_modes} lists the game mode numbers. 

\begin{table}[h!]
\centering
\caption{Plane Tank modes}
\label{table:plane_modes}
\begin{tabular}{c|cc}
    Mode & Guided Missiles & Jet plane \\ \midrule
    15 & F & F \\
    16 & T & F \\
    21 & F & T \\
    22 & T & T \\
\end{tabular}
\end{table}

\subsection{Entombed}
\label{appendix:entombed}

The original Entombed Atari game does not have an official scoring method in two player (unlike in one player). However, the official manual lists two types of gameplay, competitive play and cooperative play. We implement and distinguish between these two scoring methods using modes:

\begin{itemize}
    \item Competitive play (Mode 2, default): A player is rewarded, and their opponent penalized when the opponent loses a life.  
    \item Cooperative play (Mode 3): Similar (but not identically) to single player mode, both players are are rewarded 1 point after passing each of the 5 invisible sections of a particular stage or restarting a stage. Since restarting a stage occurs after losing a life, the players do receive a reward after dying.  
\end{itemize}

\subsection{Maze Craze}
\label{appendix:maze_craze}

Maze Craze offers numerous game types, and a visibility setting for each game type. Representing the mode as $4n+k$, the game mode is $n$ and the visibility mode is $k$. If $k=0$ the maze is fully visible, if $k=1$ or $k=2$ only part of the map is visible, and if $k=3$. 
The modes below are the fully visible versions of their game type; only ones we believe are of interest are included for brevity:

\begin{itemize}
    \item Race (mode 0): First to the end of the maze wins.
    \item Robbers (mode 4): Randomly moving robbers will kill you if you run into them. Avoid the robbers and complete the maze.
    \item Capture (mode 44): You need to capture the randomly moving robbers before you can complete the maze. The players can also place green squares that appear identical to the maze walls, confusing their opponents pathing.
\end{itemize}

\subsection{Space invaders}

Space invaders modes offer a set of 5 possible options, all combinations of which are possible.

\begin{itemize}
    \item Moving Shields: Shields move back and forth. Using them for protection is therefore more difficult. 
    \item Zigzagging Bombs: Alien's bombs randomly move horizontally, making them harder to avoid. 
    \item Fast Bombs: Alien's bombs move much faster, making them much harder to avoid. 
    \item Invisible Invaders: Aliens are invisible, making them much harder to hit.
    \item Alternating Turns: Ability to fire alternates between the players. The switch occurs either when you fire, or after a set period of time.
\end{itemize}

A particular combination of options can be set by using the following formula (where variables are encoded as 0 or 1):

$$\begin{aligned}
33 &+ \textsc{ Moving Shields} \\
&+ 2\textsc{ Zigzagging Bombs}\\
&+4 \textsc{ Fast Bombs}\\
&+8\textsc{ Invisible Invaders}\\
&+16\textsc{ Alternating Turns}
\end{aligned} $$


\subsection{Video Olympics}
\label{appendix:video_olympics}

Video Olympics is best known for containing Pong, the classic game. However, as its name suggests, Video Olympics contains a wide variety of multiplayer games. Below is a list of the games we've found to be of note for reinforcement learning research. Table \ref{table:video_olympics} contains the mode numbers. 

\begin{itemize}
    \item Classic Pong: The classic game.
    \item Quadrapong: All 4 sides of the map are protected by one player. Score is tracked by team.
    \item Volleyball: Keep the ball from hitting the ground on your side of the court. The players can "jump" vertically as well as move horizontally.
    \item Foozpong: There are multiple intertwined rows of defense and offense between the goals.
    \item Basketball: Put the ball in your opponent's hool.
\end{itemize}

\begin{table}[h!]
\centering
\caption{Video Olympics Modes}
\label{table:video_olympics}
\begin{tabular}{c|cc}
    Game & Two player & Four player \\ \midrule
    Classic Pong & 4 & 6 \\
    Foozpong   & 19 & 21 \\
    Quadrapong & N/A & 33 \\
    Volleyball & 39 & 41 \\
    Basketball & 45 & 49 \\
\end{tabular}
\end{table}


\section{Preprocessing}
\label{appendix:preprocessing}

Preprocessing was performed with SuperSuit wrappers \citep{SuperSuit2020}. To introduce non-determinism as suggested by \citet{machado2018revisiting}, sticky actions are used with a repeat probability of 25\%. Observations are generated by ALE as grayscale images, then resized to 84x84 with area interpolation. There is a frame skip of 4, followed by a frame stack of 4. Finally, an agent indicator, a 1-hot value indicating whether the current agent is the first or second player, is appended as a channel to the image. Reward clipping between -1 and 1 was used during training only.

\section{Hyperparameters}
\label{appendix:hyperparameters}

\autoref{tab:hyperparameters} Shows the hyperparameters used for training the baseline.

\begin{table}[ht]
\centering
    \begin{tabular}{l c}
    \toprule
    Hyperparameter & Value \\
    \midrule
    \texttt{adam\_epsilon} & 0.00015 \\
    \texttt{buffer\_size} & 80000 \\
    \texttt{double\_q} & \texttt{True} \\
    \texttt{dueling} & \texttt{True} \\
    \texttt{epsilon\_timesteps} & 200000 \\
    \texttt{final\_epsilon} & 0.01 \\
    \texttt{final\_prioritized\_replay\_beta} & 1.0 \\
    \texttt{gamma} & 0.99 \\
    \texttt{learning\_starts} & 80000 \\
    \texttt{lr} & 0.0001 \\
    \texttt{n\_step} & 3 \\
    \texttt{num\_atoms} & 1 \\
    \texttt{num\_envs\_per\_worker} & 8 \\
    \texttt{num\_gpus} & 1 \\
    \texttt{num\_workers} & 8 \\
    \texttt{prioritized\_replay} & \texttt{True} \\
    \texttt{prioritized\_replay\_alpha} & 0.5 \\
    \texttt{prioritized\_replay\_beta} & 0.4 \\
    \texttt{prioritized\_replay\_beta\_} & 2000000 \\
    \hspace{6mm}\texttt{annealing\_timesteps} &  \\
    \texttt{rollout\_fragment\_length} & 32 \\
    \texttt{target\_network\_update\_freq} & 50000 \\
    \texttt{timesteps\_per\_iteration} & 25000 \\
    \texttt{train\_batch\_size} & 512 \\
    \bottomrule
    \end{tabular}
    \caption{Hyperparameters for ApeX DQN on each Atari environment.}
    \label{tab:hyperparameters}
\end{table}



\end{document}